\documentclass[
twocolumn,
]{ceurart}

\usepackage{listings}
\usepackage{hyperref}

\begin{document}

\copyrightyear{2022}
\copyrightclause{Copyright for this paper by its authors.
  Use permitted under Creative Commons License Attribution 4.0
  International (CC BY 4.0).}

\conference{CDCEO 2022: 2nd Workshop on Complex Data Challenges in Earth 
Observation, July 25, 2022, Vienna, Austria}

\title{METER-ML: A Multi-Sensor Earth Observation Benchmark for Automated Methane Source Mapping}



\author[1]{Bryan Zhu}[%
email=bwzhu@cs.stanford.edu,
]
\cormark[1]
\fnmark[1]
\address[1]{Department of Computer Science, Stanford University}

\author[2]{Nicholas Lui}[%
email=niclui@stanford.edu,
]
\fnmark[1]
\address[2]{Department of Statistics, Stanford University}

\author[1]{Jeremy Irvin}[%
email=jirvin16@cs.stanford.edu,
]
\fnmark[1]

\author[1]{Jimmy Le}[%
email=jimmyle@cs.stanford.edu,
]

\author[3]{Sahil Tadwalkar}[%
email=stadwalk@stanford.edu,
]
\address[3]{Department of Civil and Environmental Engineering, Stanford University}

\author[4]{Chenghao Wang}[%
email=chenghao.wang@stanford.edu,
]

\author[4]{Zutao Ouyang}[%
email=ouyangzt@stanford.edu,
]

\author[4]{Frankie Y. Liu}[%
email=frankliu@stanford.edu
]
\address[4]{Department of Earth System Science, Stanford University}

\author[1]{Andrew Y. Ng}[%
email=ang@cs.stanford.edu,
]

\author[4,5]{Robert B. Jackson}[%
email=Rob.Jackson@stanford.edu,
]
\address[5]{Woods Institute for the Environment and Precourt Institute for Energy, Stanford University}

\cortext[1]{Corresponding author.}
\fntext[1]{These authors contributed equally.}

\begin{abstract}
Reducing methane emissions is essential for mitigating global warming. To attribute methane emissions to their sources, a comprehensive dataset of methane source infrastructure is necessary. Recent advancements with deep learning on remotely sensed imagery have the potential to identify the locations and characteristics of methane sources, but there is a substantial lack of publicly available data to enable machine learning researchers and practitioners to build automated mapping approaches. To help fill this gap, we construct a multi-sensor dataset called METER-ML containing 86,599 georeferenced NAIP, Sentinel-1, and Sentinel-2 images in the U.S. labeled for the presence or absence of methane source facilities including concentrated animal feeding operations, coal mines, landfills, natural gas processing plants, oil refineries and petroleum terminals, and wastewater treatment plants. We experiment with a variety of models that leverage different spatial resolutions, spatial footprints, image products, and spectral bands. We find that our best model achieves an area under the precision recall curve of 0.915 for identifying concentrated animal feeding operations and 0.821 for oil refineries and petroleum terminals on an expert-labeled test set, suggesting the potential for large-scale mapping.
We make METER-ML freely available at \href{https://stanfordmlgroup.github.io/projects/meter-ml}{this link} to support future work on automated methane source mapping.
\end{abstract}


\maketitle

\section{Introduction}

\begin{figure}
    \centering
    \caption{METER-ML is a multi-sensor dataset containing 86,599 examples of NAIP aerial imagery, Sentinel-2 satellite imagery, and Sentinel-1 satellite imagery. We include 19 spectral bands across these three products, with the RGB and VH\&VV bands shown here. Each example is labeled with the presence or absence of six different methane source facilities and is georeferenced. A small amount of examples are labeled to contain facilities from more than one category and 34,870 examples contain no facilities from the six categories.}
    \includegraphics[width=0.5\textwidth]{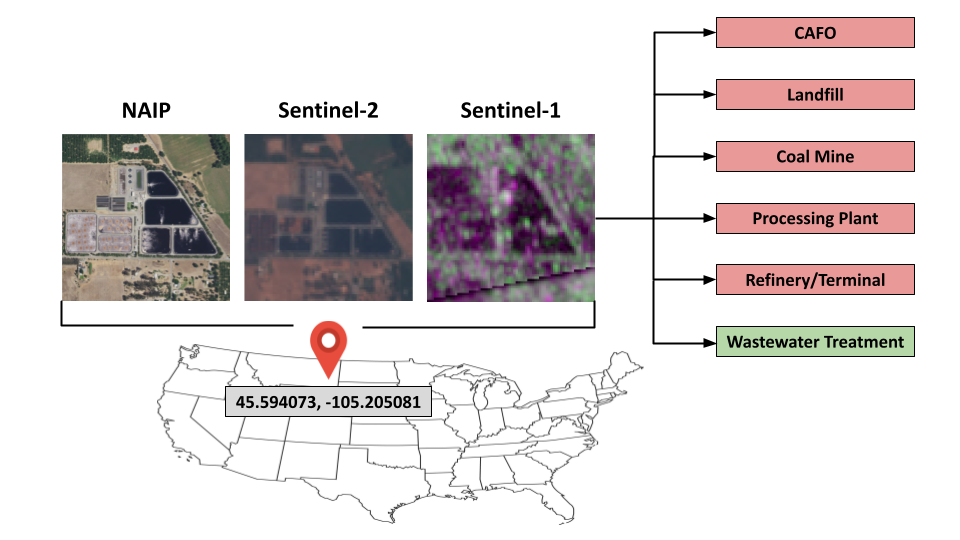}
    \label{fig:main}
\end{figure}

Anthropogenic methane emissions are the main contributor to the rise of atmospheric methane \cite{zhang2022anthropogenic}, and mitigating methane emissions is widely recognized as crucial for slowing global warming and achieving the goals of the Paris Agreement \cite{agreement2015paris}. Multiple satellites are in orbit or launching soon which will measure methane emissions from the surface using top-down approaches, but in order to attribute these emissions to specific sources on the ground, a comprehensive database of methane emitting infrastructure is necessary \cite{jacob2022quantifying}. Although several public databases of this infrastructure exist, the data available globally is incomplete, erroneous, and unaggregated. 

AI approaches on Earth observation data have the potential to fill in this gap. Several recent works have developed deep learning models to automatically interpret remotely sensed imagery and deploy them at scale to map infrastructure \cite{yu2018deepsolar,lee2021scalable,kruitwagen2021global,sirko2021continental}. Methods for mapping methane source infrastructure have been emerging as well, including well pads in the Denver basin \cite{dileepautomated}, oil refineries and concentrated animal feed operations in the U.S. \cite{sheng2020ognet,handan2019deep}, and wastewater treatment plants in Germany \cite{li2022leveraging}. Each of these works depended on the curation of large, labeled datasets to develop the machine learning models, but there is a lack of publicly available, labeled Earth observation data, specifically on methane emitting infrastructure, which prohibits researchers and practitioners from building automated mapping approaches.

In this work, we construct a multi-sensor Earth observation dataset for methane source infrastructure identification called METER-ML. In support of a new initiative to build a global database of methane emitting infrastructure called the \textbf{ME}thane \textbf{T}racking \textbf{E}missions \textbf{R}eference (METER) \cite{meter}, we develop METER-ML to allow the machine learning community to experiment with multi-view/multi-modal modeling approaches to automatically identify this infrastructure in remotely sensed imagery. METER-ML includes georeferenced imagery from three remotely sensed image products, specifically 19 spectral bands in total from NAIP, Sentinel-1, and Sentinel-2, capturing 51,729 sources of methane from six different classes as well as 34,870 negative examples (Figure~\ref{fig:main}). The dataset includes expert-reviewed validation and test sets for robustly evaluating the performance of derived models. Using the dataset, we experiment with a variety of convolutional neural network models which leverage different spatial resolutions, spatial footprints, image products, and spectral bands. The dataset is freely available\footnote{https://zenodo.org/record/6911013} in order to encourage further work on developing and validating methane source mapping approaches.

\section{Methods}
\subsection{Methane source locations}
We collect locations of methane emitting infrastructure in the U.S. from a variety of public datasets. We focus on the U.S. in this study due to the high availability of publicly accessible infrastructure data and remotely sensed imagery. The infrastructure categories we include are concentrated animal feeding operations (CAFOs), coal mines (Mines), landfills (Landfills), natural gas processing plants (Proc Plants), oil refineries and petroleum terminals (including crude oil and liquified natural gas terminals), and wastewater treatment plants (WWTPs). We group oil refineries and petroleum terminals together due to their high similarity in appearance, and refer to that category as ``Refineries \& Terminals'' (R\&Ts). These infrastructure categories were chosen based on their potential for emitting methane along with their consistent, visible differentiating features which make them feasible to identify in high resolution remotely sensed imagery.

The locations are obtained from 18 different publicly available datasets, all of which have licenses that allow for redistribution (see Table~\ref{tab:sources} in the Appendix). As various datasets may contain the same locations of infrastructure, we deduplicate by considering locations within 500m of each other identical. In total we include 51,729 unique locations of methane source infrastructure in the dataset, which we refer to as positive examples.

\begin{table}
\centering
\caption{Counts and proportions of each category in METER-ML. The labels on the training set are obtained from public data whereas the labels on validation and test sets are obtained from a consensus of two methane source identification experts. The individual category counts do not add up to the overall train/valid/test counts as some (0.8\%) of the positive examples are labeled with more than one methane source category.}
\setlength{\tabcolsep}{2pt}
\resizebox{\columnwidth}{!}{
\begin{tabular}{c|ccc|c}
    Category & Train (\%) & Valid (\%) & Test (\%) & Total \\
    \hline
    CAFOs & 24957 (29.3\%) & 47 (9.1\%) & 92 (9.0\%) & 25096 \\
    Landfills & 4085 (4.8\%) & 46 (8.9\%) & 111 (10.9\%) & 4242 \\
    Coal Mines & 1776 (2.1\%) & 40 (7.8\%) & 72 (7.1\%) & 1888 \\
    Proc Plants & 1900 (2.2\%) & 38 (7.4\%) & 107 (10.5\%) & 2045 \\
    R\&Ts & 4012 (4.7\%) & 59 (11.5\%) & 108 (10.6\%) & 4179 \\
    WWTPs & 14519 (17.1\%) & 46 (8.9\%) & 129 (12.7\%) & 14694 \\
    Negatives & 34195 (40.2\%) & 249 (48.3\%) & 426 (41.8\%) & 34870 \\
    \hline 
    Total & 85066 & 515 & 1018 & 86599
\end{tabular}
}
\label{tab:composition}
\end{table}

\subsection{Negative locations}
We additionally include a variety of images in the dataset which capture none of the six methane emitting facilities. To do this, we define around 50 classes (see Appendix) of different facilities and landscapes and select characteristic examples of each class. Then we collect locations containing similar facilities and landscapes using the Descartes Labs GeoVisual Search \cite{keisler2019visual}, providing up to 1000 similar locations per example. A sample of the similar locations were manually vetted in each case to ensure no locations obtained actually corresponded to the six methane source categories. In total we include 34,870 locations of facilities and landscapes which are not any of the six infrastructure categories, and refer to these as negative examples. The counts and proportions of the positive and negative classes in the dataset are shown in Table~\ref{tab:composition}.

\subsection{Remotely sensed imagery}
We pair all of the locations in the dataset with three publicly available remotely sensed image sources. Specifically we include aerial imagery from the USDA National Agriculture Imagery Program (NAIP) as well as satellite imagery captured by Sentinel-1 (S1) and Sentinel-2 (S2). NAIP imagery covers the contiguous U.S. and S1 and S2 imagery both have global coverage. For NAIP we use 1m resolution imagery, for Sentinel-2 we use the L1C product at 10m resolution, and for Sentinel-1 we use the Sigma Nought Backscatter product at 10m resolution. We use all spectral bands from each product. Specifically, we use the three visible (RGB) and single near-infrared (NIR) bands from NAIP and S2, the single coastal aerosol (CA) band, four red-edge (RE1-4) bands, single water vapor (WV) band, single cirrus (C) band, and the two shortwave infrared (SWIR1-2) bands from S2, and the V-transmit (VH and VV) bands from S1. We include S1 and S2 in the dataset in order to enable experimenting with coarser resolution satellite imagery which is globally available, unlike NAIP. The details of each imagery product and band are shown in Table \ref{tab:imagery}.

In order to construct images containing each location in the dataset, we consider a 720m x 720m footprint centered around the location. This footprint was chosen to balance the size of the images with the contextual information, but we investigate this choice in the experiments. Due to the geographic coordinate noise in the publicly available datasets, we chose to center the imagery at the locations which increases the likelihood the facilities are captured in the imagery, but still has natural variation in the locations of the facilities in the imagery. We construct a mosaic of the most recently captured pixels in a time range for each image product, where we consider NAIP images captured between 2017 and 2021 and Sentinel-1 and Sentinel-2 images between May and September 2021, where Sentinel-2 images are selected based on lowest cloud cover. We use the Descartes Labs platform to download all of the imagery \cite{dl}.

The total dataset contains 86,599 images capturing ten spectral bands across the three imagery products. Information about the remotely sensed image products and bands included in the dataset are provided in Table~\ref{tab:imagery} and characteristic examples for each methane source category are shown in Figure~\ref{fig:examples} in the Appendix.

\begin{table}
\caption{Summary of the remotely sensed image products and bands included in METER-ML. RGB are the three visible bands, NIR is a single near-infrared band, RE1-4 are the four red-edge bands, SWIR1-2 are the two shortwave infrared bands, CA is the single coastal aerosol band, WV is the single water-vapor band, C is the single cirrus band, and VH \& VV are the two V-transmit bands.}
\setlength{\tabcolsep}{2pt}
\resizebox{\columnwidth}{!}{
\begin{tabular}{ccccc}
    Product & Bands & Image Size & Resolution \\
    \hline
    NAIP & RGB \& NIR & 720x720 & 1m \\
    Sentinel-2 & RGB \& NIR & 72x72 & 10m \\
    Sentinel-2 & RE1-4 \& SWIR1-2 & 36x36 & 20m \\
    Sentinel-2 & CA \& WV \& C & 12x12 & 60m \\
    Sentinel-1 & VH \& VV & 72x72 & 10m
\end{tabular}
}
\label{tab:imagery}
\end{table}


\subsection{Validation and test sets}
Two Stanford University postdoctoral researchers with expertise in methane emissions and related infrastructure individually reviewed 1,533 examples to compose the held-out validation and test sets. To determine which examples to include in these held-out sets, we randomly sampled 150 images from each of the six positive classes as well as a random sample of 33 images which have multiple labels, constituting 933 positive examples according to the original public dataset labels. We additionally sampled 12 images from each of the 50 negative categories resulting in 600 negative examples. The experts both manually reviewed these examples and identified the presence or absence of the six methane source categories by using a combination of NAIP imagery as well as Google Maps imagery, which often had finer spatial resolution as well as place names. The facility had to be captured by the NAIP image for the corresponding label to be assigned. If the expert identified no clearly visible methane source categories in the image, the example was labeled ``negative'', and if the expert was uncertain about any label, the example was labeled ``uncertain''. The two labels per example were then resolved as follows:

\begin{enumerate}
    \item If the experts agreed and neither was uncertain, the agreed upon label was taken as the final label.
    \item If the experts disagreed, and one was uncertain but the other was not, the expert's certain label was taken as the final label.
    \item If the experts disagreed, but one agreed with the original label, the original label was taken as the final label.
    \item In all other scenarios, the example was reviewed jointly by the experts and a final label was assigned.
\end{enumerate}

Only 76 examples out of the 1,533 went to another round of review. The resulting datasets have 858 positive examples and 675 negative examples. We split the 1,533 examples into 515 for the validation set and 1,018 for the test set. The label counts on the validation and test sets are shown in Table~\ref{tab:composition}.

\section{Experiments}
We run a variety of multi-label classification experiments on the curated dataset. In all of our experiments, we use a DenseNet-121 convolutional neural network architecture \cite{huang2017densely}. Preliminary experiments on the dataset explored various ResNet and DenseNet architectures and found that DenseNets outperformed all ResNet variants \cite{he2016deep}. We use a linear layer which outputs six values indicating the likelihood that each of the six methane source categories are present in the input image, which outperformed individual models across all classes in our preliminary experiments. Although the model does not explicitly produce a value indicating the likelihood that the image is negative, a low value assigned to all classes indicates a negative prediction. The loss function is the mean of six unweighted binary cross entropy losses, where the label is 1 if the class if present in the image and 0 otherwise. All six labels in the negative examples are 0. The network weights are initialized with weights from a network pre-trained on ImageNet \cite{deng2009imagenet}. Before inputting the images into the networks, we upscale the Sentinel-1 and Sentinel-2 images to match the size of NAIP images using bilinear resampling and normalize the values by the display range of the bands (see Table~\ref{tab:bands} in the Appendix). When using inputs with less than or more than 3 channels, we replace the first convolutional neural network layer with one which accepts the corresponding number of channels. Each model is trained for 5 epochs with a batch size of 4. For each model we use the checkpoint saved after an epoch which led to the lowest validation loss. We use an Adam optimizer with standard parameters \cite{kingma2014adam} and a learning rate of 0.02. All models are trained using a GeForce GTX 1070 GPU.

The baseline setting for all experiments uses images capturing a footprint of 720m x 720m with 1m spatial resolution (720 x 720 image dimensions). After the models are trained, each of the six values output by the model are fed through an element-wise sigmoid function to produce a probability for each of the six categories. To evaluate the performance of the models, we compute the per-class area under the precision recall curve (AUPRC) and summarize the performance over all classes by taking the macro-average of the per-class AUPRCs.

\begin{table*}[t]\centering
\caption{Per-class and overall (macro-average) validation AUPRC for different remotely sensed image products and bands. All of these experiments use images of size 720x720 at a spatial resolution of 1m per pixel, with S1 and S2 upsampled to that resolution.}
\setlength{\tabcolsep}{3pt}
\begin{tabular}{cc|cccccc|c}
    Image Product & Bands & CAFOs & Landfills & Mines & Proc Plants & R\&Ts & WWTPs &  Overall \\
    \hline
    S1 & VH\&VV & 0.519 & 0.107 & 0.152 & 0.218 & 0.487 & 0.119 & 0.267\\
    S2 & RGB & 0.889 & 0.268 & 0.305 & 0.374 & 0.694 & 0.204 & 0.456 \\
    S2 & All & 0.889 & 0.189 & 0.382 & 0.368 & 0.690 & 0.183  & 0.450\\
    S2 \& S1 & All & 0.923 & 0.152 & 0.379 & 0.391 & 0.612 & 0.231 & 0.448\\
    NAIP & RGB & 0.903 & 0.270 & 0.348 & 0.327 & 0.849 &  0.182 & 0.480 \\
    NAIP & All & \textbf{0.945} & \textbf{0.276} & 0.401 & \textbf{0.508} & \textbf{0.857} & \textbf{0.303} & \textbf{0.548}\\
    NAIP \& S2 \& S1 & All & 0.889 & 0.214 & \textbf{0.473} & 0.457 & 0.796 & 0.272 & 0.517
\end{tabular}
\label{tab:product}
\end{table*}

\subsection{Impact of using different imaging products and bands}
We investigate the impact of using different combinations of image products and bands in the dataset (Table~\ref{tab:product}). Specifically, we experiment with NAIP, S2, and S1 alone, only visible bands and all spectral bands for S2 and NAIP, all spectral bands from S1 and S2 together (representing the model closest to public global transferability due to the global coverage of S1 and S2), and all spectral bands from the three products together.

The best model according to macro-average AUPRC is the one which uses NAIP with  all bands (the three visible bands and NIR band), achieving an overall AUPRC of 0.548 and the highest performance on CAFOs, Landfills, Proc Plants, R\&Ts, and WWTPs compared to all other tested product and band combinations. Notably, it achieves very high performance on CAFOs (AUPRC=0.945) and high performance on R\&Ts (AUPRC=0.857). The second best model is the joint NAIP+S2+S1 model, achieving an overall AUPRC of 0.517 and the highest performance on Mines (AUPRC=0.473) compared to all other tested product and band combinations.

S1 alone underperforms all other combinations of products and bands, followed by S2 and S1 jointly, which performed similarly overall to S2 with only the visible bands and all spectral bands. Importantly, the S2 and S1 joint model still achieves high performance on CAFOs (AUPRC=0.923), although the performance is lower than performance on CAFOs using NAIP imagery (AUPRC=0.947). There is a significant drop in performance on all classes when moving from NAIP to S2, highlighting the benefit of using high spatial resolution imagery.

The inclusion of the non-visible information substantially improves overall AUPRC for NAIP (AUPRC=0.480 $\rightarrow$ 0.548) but not for Sentinel 2 (AUPRC=0.450 $\rightarrow$ 0.448). For NAIP, the improvement is observed for all classes, with substantial gains on CAFOs, Mines, Proc Plants, and WWTPs. For Sentinel 2, the inclusion of non-visible bands substantially improves performance on Mines but substantially degrades performance on Landfills. For both products, minimal change on R\&Ts performance is observed when including the non-visible bands.

\begin{table*}[t]\centering
\caption{Per-class and overall (macro-average) validation AUPRC at varying image footprints and spatial resolutions.}
\setlength{\tabcolsep}{3pt}
\begin{tabular}{cc|cccccc|c}
    Image Footprint & Resolution & CAFOs & Landfills & Mines & Proc Plants & R\&Ts & WWTPs &  Overall \\
    \hline
    240x240 & 1m & 0.773 & 0.217 & 0.407 & 0.438 & 0.735 & 0.337 & 0.485 \\
    480x480 & 1m & 0.772 & 0.226 & 0.260 & 0.371 & 0.855 & \textbf{0.506} & 0.498\\
    720x720 & 3m & 0.891 & 0.245 & 0.378 & \textbf{0.566} & 0.837 & 0.269 & 0.531\\
    720x720 & 1.5m & 0.927 & 0.244 & \textbf{0.426} & 0.366 & 0.831 & 0.449 & 0.541\\
    720x720 & 1m & \textbf{0.945} & \textbf{0.276} & 0.401 & 0.508 & \textbf{0.857} & 0.303 & \textbf{0.548} \\
\end{tabular}
\label{tab:footres}
\end{table*}

\begin{table}[t]\centering
\caption{Per-class and overall (macros-average) test metrics of the per-class expert model. The per-class expert model consists of one model per class, where the model used for each class is selected based on the highest performing settings for that class across the product, bands, footprint, and resolution experiments.}
\setlength{\tabcolsep}{2pt}
\begin{tabular}{cccccc}
Category & AUPRC & AUROCC & Precision & Recall & F1\\
\hline 
    CAFOs & 0.915 & 0.989 & 0.822 & 0.902 & 0.860\\ 
    Landfills & 0.259 & 0.754  & 0.246  & 0.523 &  0.334\\
    Mines & 0.470 &  0.905 & 0.558 & 0.403 & 0.468 \\
    Proc Plants & 0.350 & 0.787 & 0.336 & 0.477  & 0.394 \\
    R\&Ts & 0.821 & 0.956 & 0.752 & 0.787 & 0.769\\
    WWTPs & 0.534 & 0.836 & 0.633 & 0.477 & 0.544 \\
    \hline
    Overall & 0.558  & 0.871  & 0.558  & 0.595 & 0.562
\end{tabular}
\label{tab:test}
\end{table}

\subsection{Impact of image footprint and spatial resolution}
As image footprint (i.e. the amount of area on the ground captured by the image) and spatial resolution likely impact model performance due to the variation in the sizes of the methane-emitting facilities and equipment, we conduct experiments to test these effects (Table~\ref{tab:footres}). To investigate the impact of footprint, we center crop the 720 x 720 1m images to obtain 480 x 480 and 240 x 240 1m images corresponding to 480m x 480m and 240m x 240m footprints respectively. Note that this reduces the area on the ground with spatial resolution held constant. To investigate the impact of spatial resolution, we use cubic spline interpolation \cite{parsania2016comparative} to downsample the 720 x 720 images to 480 x 480 (1.5m resolution, corresponding to Airbus SPOT imagery) and 240 x 240 (3m resolution, corresponding to PlanetScope imagery). Note that this reduces the spatial resolution without modifying the image footprint. In all experiments, we up-sample the images back to 720 x 720 to avoid any differences in performance due to varying image size. We use NAIP with RGB + NIR bands for these experiments as this setting produced the best overall performance compared to the other combinations of products and bands.


We find that the largest tested image footprint achieves the highest overall performance (0.548) and substantially outperforms both smaller spatial footprints across all classes except for WWTPs. This may be explained by the fact that a significant number of smaller wastewater treatment plants are surrounded by industrial buildings and other infrastructure, so cropping out this infrastructure improves the model's ability to identify the salient features of the wastewater treatment facilities.

We further find that the highest spatial resolution achieves the best overall performance (AUPRC=0.548), outperforming the coarser resolution models on CAFOs, Landfills, and R\&Ts. The 1.5m resolution model closely follows with an overall AUPRC of 0.541 and outperforms the 1m resolution model on Mines. The 3m resolution model also closely follows the 1.5m resolution model achieving an overall performance of 0.531, and substantially outperforms both the higher resolution models on Proc Plants. This result suggests that models developed at 1.5m and even 3m resolution have the potential to perform almost as well as 1m resolution models, which has implications on global applicability as Airbus SPOT and PlanetScope are globally (privately) available at 1.5m and 3m resolution respectively.


\subsection{Per-class expert model test set results}
For each methane source category, we select the experimental configuration (product/band/footprint/resolution) that achieved the highest validation AUPRC for that class to serve as the ``class expert''. We refer to the combination of the different class experts as the per-class expert model.

We evaluate the per-class expert model on the hold-out test set using a variety of metrics including AUPRC and area under the receiver operating characteristic curve (AUROCC) as well as precision, recall, and F1 at the threshold which achieves the highest F1 on the validation set. The results are shown in Table~\ref{tab:test}. The per-class expert model obtains a macro-average AUPRC of 0.558. The model does especially well on CAFOs (AUPRC=0.915) and R\&Ts (AUPRC=0.821), possibly because these sources have very distinctive features (e.g., long barns in CAFOs and storage tank farms in R\&Ts). It performs more poorly on the other sources, especially landfills which do not have many clear distinctive features visible at 1m resolution. Notably it achieves the lowest performance on the categories with the least number of examples in the dataset, excluding R\&Ts which may be simpler to identify due to their homogeneity and discernible features.

\section{Discussion}
The experiments suggest that the choice of imaging product, spectral band, image footprint, and spatial resolution can lead to substantial differences in model performance, with the effect often depending on the methane source category. In particular, this suggests that there is significant room to explore approaches which leverage the multi-sensor and multi-spectral aspects of METER-ML. For example, the NAIP \& S2 \& S1 model underperformed the model which used NAIP alone, and using all 13 spectral bands in the S2 model did not lead to substantial performance differences compared to the S2 model which only used the three visible bands. We also do not leverage the geographic information explicitly in the models, but this has been shown to improve performance on other Earth observation tasks \cite{mac2019presence,irvin2020forestnet}. Furthermore, there is potential to augment the dataset with other sources of imagery and information available at the provided geographic locations. We hope to help create new versions of METER-ML which may include other sources of input data and methane emitting infrastructure categories.

The best model from our experiments achieves high performance on identifying CAFOs and R\&Ts, suggesting the potential to map these facilities with NAIP imagery in the U.S. which aligns with findings from prior studies \cite{sheng2020ognet,handan2019deep}. The performance for identifying CAFOs remains high when using S1 and S2, which are globally and publicly available. This suggests the potential to use these lower spatial resolution imagery sources to map CAFOs in other countries besides the U.S., but future work should investigate whether these findings generalize to other regions. There is still a large gap to achieving high performance for each of the other methane source categories and further improve performance on the high performing categories, so METER-ML is a challenging benchmark to test new infrastructure identification approaches.

There are many other publicly available remote sensing datasets for classification, with some of the most common being UC Merced \cite{yang2012geographic}, SAT-4 and SAT-6 \cite{basu2015deepsat}, AID \cite{xia2017aid}, NWPU-RESISC45 \cite{cheng2017remote}, EuroSAT \cite{helber2019eurosat}, and  BigEarthNet \cite{sumbul2019bigearthnet}. Few of these datasets have georeferenced multi-sensor images, which limits their utility for new modeling approaches and downstream use. The OGNet dataset \cite{sheng2020ognet} is the most similar publicly available dataset to METER-ML and is essentially a subset of it, containing NAIP imagery of refineries in the contiguous U.S.

We identify four limitations of this work. First, we limit the geographic scope of METER-ML to the U.S. due to the availability of disseminatable infrastructure data and publicly available, high resolution imagery. Future work should include data in other regions worldwide. Second, we do not include longitudinal imagery in the dataset to reduce the size and complexity of the dataset as most infrastructure is static over time. However, longitudinal information has the potential to provide additional signal to help differentiate certain facilities, e.g. waste pile evolution at landfills. Third, we use a DenseNet121 model that is pre-trained on ImageNet, but the shape and number of channels of remote sensing imagery can be significantly different from ImageNet. It would be worthwhile to train a network from scratch on METER-ML, and compare its performance against a network that is pre-trained on ImageNet and fine-tuned on METER-ML. Fourth, our approach to combine the multi-sensor data may not be optimal as the products and spectral bands have different spatial resolutions and sensor types (e.g. active vs. passive sensors). One alternate approach may be to dedicate different network branches for the inputs and combine the representations from each branch.

\section{Conclusion}
In this work, we curate a large georeferenced multi-sensor dataset called METER-ML to test automated methane source identification approaches. We conduct a variety of experiments investigating the impact of remotely sensed image product, spectral bands, image footprint, and spatial resolution on model performance measured against a consensus of expert labels. We find that a model which leverages NAIP with all four bands achieves the highest overall performance across the tested image product and spectral band combinations, followed closely by a joint NAIP, Sentinel-2, and Sentinel-1 model. We also find that the highest spatial resolution and footprint leads to the best overall performance, although performance can depend on the methane source category. Finally we show that the best model achieves high performance in identifying concentrated animal feeding operations and oil refineries and petroleum terminals, suggesting the potential to map them at scale, but substantially lower performance on the other four categories with notably lower performance identifying processing plants and landfills. We make METER-ML freely available in order to encourage and support future work on developing Earth observation models for mitigating climate change.

\begin{acknowledgments}
This work was supported by the High Tide Foundation to construct the METER database. We acknowledge Rose Rustowicz and Kyle Story for their support of this work, as well as the Descartes Labs Platform API and tools for downloading and processing the remotely sensed imagery. We also thank Ritesh Gautam and Mark Omara for their help working with the oil and gas infrastructure data, Evan Sherwin for his advice on the dataset and methane source categories, and Victor Maus for providing the coal mines data.
\end{acknowledgments}

\bibliography{sample-ceur}

\appendix
\onecolumn
\section{Appendix}
\subsection{Methane-Emitting Infrastructure Datasets}\vspace{-0.5cm}
\begin{table*}[h]\centering
\caption{Summary of the datasets containing the locations of methane-emitting infrastructure that are included in METER-ML. All datasets are subsetted to the locations within the contiguous U.S. We use the centroids of polygons for any datasets provided in polygon format.}
\resizebox{0.7\columnwidth}{!}{
\begin{tabular}{cccc}
    Dataset Source & Scope & Methane Source Categories \\
    \hline
    CA Energy Commission \cite{caloes} & California & R\&Ts \\
    CSWRCB \cite{cswrcb} & California & WWTPs \\
    data.world \cite{data_world} & Michigan & CAFOs  \\
    Data For Cause Challenge \cite{dataforcause} & US & CAFOs  \\
    EIA \cite{eia} & US & Proc Plants, R\&Ts, WWTPs  \\
    GHGRP \cite{ghgrp} & US & Landfills, Mines, R\&Ts, WWTPs  \\
    GOGI \cite{gogi} & Global & Proc Plants, R\&Ts \\
    HIFLD \cite{hifld} & US & R\&Ts, WWTPs  \\
    HydroWASTE \cite{hydrowaste} & Global & WWTPs \\
    IndianaMap \cite{indianamap} & Indiana & CAFOs \\
    LMOP \cite{lmop} & US & Landfills \\
    Marchese et al. \cite{marchese2015methane} & US & Proc Plants  \\
    Maus et al. \cite{maus2022update} & Global & Mines  \\
    Minnesota Metropolitan Council \cite{metro_wwtp} & Minnesota & WWTPs \\
    Minnesota Pollution Control Agency \cite{mpca} & Minnesota & CAFOs \\
    ORNL DAAC \cite{hopkins2019sources} & California & CAFOs  \\
    Sierra Club \cite{sierra} & Michigan & CAFOs \\
    Stanford RegLab \cite{reglab} & North Carolina & CAFOs
\end{tabular}
}
\label{tab:sources}
\end{table*}

\vspace{-0.7cm}\subsubsection{Coal Mines Data}\vspace{-0.1cm}
The mines data from \cite{maus2022update} were subsetted to coal mines in order to capture the mines responsible for the vast majority of methane emissions. To do this, the polygons and
coal mine coordinates obtained from S\&P Global Commodity Insights were matched to determine which polygons were spatially related to coal mine coordinates. Then a visual check and hand cleaning was performed on the polygons assigned a coal mining label to ensure correctness.

\subsection{Negative Classes}\vspace{-0.1cm}
We identify a variety of infrastructure which are not any of the six infrastructure categories to use as negatives in the dataset. Specifically we include football fields, marinas, solar panels, large bodies of water, parking lots, windmills, baseball fields, airport runways, clouds, neighborhoods, golf courses, roundabouts, mountainous terrain, trees, boats, islands, rocks, rivers, roads, bridges, ripples in water, snow, canyon formations, sparse forests, suburban neighborhoods, beaches, clear water, swimming pools, sand, corn farms, soy farms, trees on mountainside, farm houses, grass, airplanes, turning roads, intersections, multifamily residential facilities, rapids, docks, highway loops, mowed grass, container yards, soccer fields, greenhouses, crops, personal watercrafts, pivot irrigation systems, and concrete plants. Characteristic examples of each type were selected and a variety of similar examples per type were obtained using the Descartes Labs GeoVisual Similarity tool \cite{keisler2019visual}.

\subsection{Remotely Sensed Image Statistics and Examples}\vspace{-0.5cm}
\begin{table*}[h!]\centering
\caption{Summary of the remotely sensed image products and bands included in METER-ML. The raw image data contains values from the data range, while the display range is used to normalize values before displaying imagery or inputting into models.}
\resizebox{0.6\columnwidth}{!}{
\begin{tabular}{c|cccc}
    Product/Bands & Image Size & Resolution & Data Range & Display Range\\
    \hline
    NAIP RGB \& NIR & 720x720 & 1m & [0,255] & [0,255] \\
    Sentinel-2 CA & 12x12 & 60m & [0,10000] & [0,10000] \\
    Sentinel-2 RGB & 72x72 & 10m & [0,10000] & [0,4000] \\
    Sentinel-2 RE1-4 & 36x36 & 20m  & [0,10000] & [0,10000] \\
    Sentinel-2 NIR & 72x72 & 10m & [0,10000] & [0,10000] \\
    Sentinel-2 WV & 12x12 & 60m  & [0,10000] & [0,10000] \\
    Sentinel-2 C & 12x12 & 60m  & [0,10000] & [0,10000] \\
    Sentinel-2 SWIR1-2 & 36x36 & 20m  & [0,10000] & [0,10000] \\
    Sentinel-1 VH & 72x72 & 10m & [1,4095] & [585,2100] \\
    Sentinel-1 VV & 72x72 & 10m & [1,4095] & [585,2926]
\end{tabular}
}
\label{tab:bands}
\end{table*}

\begin{figure*}[ht]
\setlength{\tabcolsep}{2pt}
    \centering
    \caption{Characteristic example images of each category in METER-ML. METER-ML contains 19 spectral bands across 3 image products, visualized here in grayscale for single bands and false color composites for multiple bands (the first listed band is used in the red channel, second in the green, third in the blue).}
    \setlength{\extrarowheight}{32pt}
    \begin{tabular}{lccc}
    Category & NAIP RGB & NAIP NIR & S1 VV\&VH \\
    \hline
        CAFOs & \raisebox{-.5\height}{\includegraphics[width=0.15\textwidth]{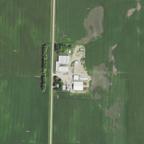}} & \raisebox{-.5\height}{\includegraphics[width=0.15\textwidth]{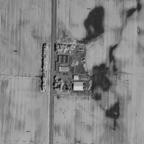}} & \raisebox{-.5\height}{\includegraphics[width=0.15\textwidth]{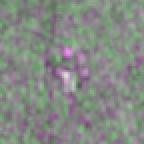}} \\
        Coal Mines & \raisebox{-.5\height}{\includegraphics[width=0.15\textwidth]{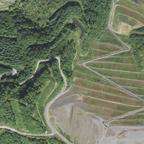}} & \raisebox{-.5\height}{\includegraphics[width=0.15\textwidth]{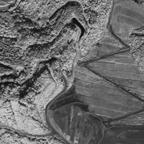}} & \raisebox{-.5\height}{\includegraphics[width=0.15\textwidth]{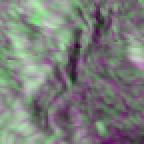}} \\
        Landfills & \raisebox{-.5\height}{\includegraphics[width=0.15\textwidth]{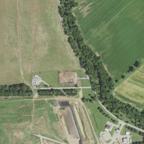}} & \raisebox{-.5\height}{\includegraphics[width=0.15\textwidth]{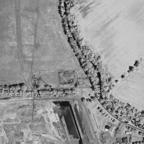}} & \raisebox{-.5\height}{\includegraphics[width=0.15\textwidth]{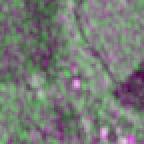}} \\
        Proc Plants & \raisebox{-.5\height}{\includegraphics[width=0.15\textwidth]{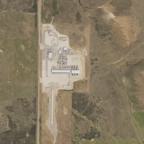}} & \raisebox{-.5\height}{\includegraphics[width=0.15\textwidth]{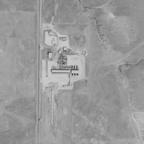}} & \raisebox{-.5\height}{\includegraphics[width=0.15\textwidth]{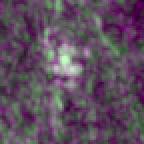}} \\
        R\&Ts & \raisebox{-.5\height}{\includegraphics[width=0.15\textwidth]{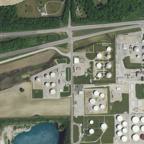}} & \raisebox{-.5\height}{\includegraphics[width=0.15\textwidth]{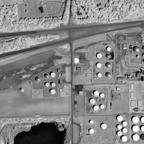}} & \raisebox{-.5\height}{\includegraphics[width=0.15\textwidth]{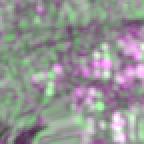}} \\
        WWTPs & \raisebox{-.5\height}{\includegraphics[width=0.15\textwidth]{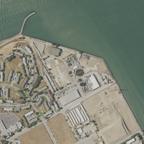}} & \raisebox{-.5\height}{\includegraphics[width=0.15\textwidth]{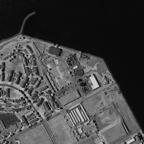}} & \raisebox{-.5\height}{\includegraphics[width=0.15\textwidth]{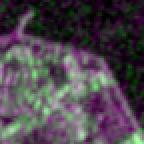}} \\
        Negatives & \raisebox{-.5\height}{\includegraphics[width=0.15\textwidth]{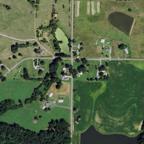}} & \raisebox{-.5\height}{\includegraphics[width=0.15\textwidth]{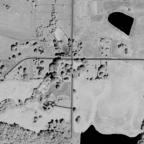}} & \raisebox{-.5\height}{\includegraphics[width=0.15\textwidth]{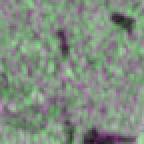}} \\
    \end{tabular}
    \label{fig:examples}
\end{figure*}

\begin{figure*}[ht]
\setlength{\tabcolsep}{2pt}
    \centering
    \setlength{\extrarowheight}{32pt}
    \begin{tabular}{lccccc}
    Category & S2 RGB & S2 NIR & S2 RE1\&SWIR1-2 & S2 RE2-4 & S2 CA\&WV\&C \\
    \hline
        CAFOs & \raisebox{-.5\height}{\includegraphics[width=0.15\textwidth]{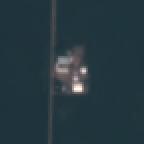}} & \raisebox{-.5\height}{\includegraphics[width=0.15\textwidth]{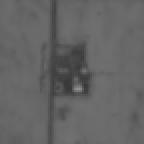}} & \raisebox{-.5\height}{\includegraphics[width=0.15\textwidth]{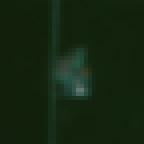}} & \raisebox{-.5\height}{\includegraphics[width=0.15\textwidth]{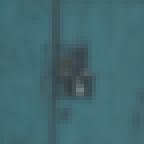}} & \raisebox{-.5\height}{\includegraphics[width=0.15\textwidth]{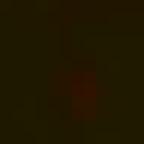}} \\
        Coal Mines & \raisebox{-.5\height}{\includegraphics[width=0.15\textwidth]{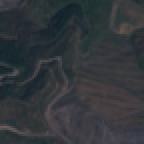}} & \raisebox{-.5\height}{\includegraphics[width=0.15\textwidth]{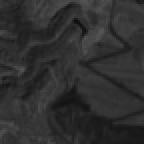}} & \raisebox{-.5\height}{\includegraphics[width=0.15\textwidth]{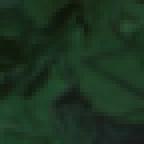}} & \raisebox{-.5\height}{\includegraphics[width=0.15\textwidth]{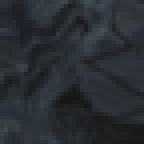}} & \raisebox{-.5\height}{\includegraphics[width=0.15\textwidth]{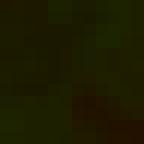}} \\
        Landfills & \raisebox{-.5\height}{\includegraphics[width=0.15\textwidth]{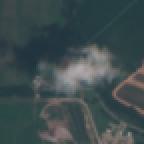}} & \raisebox{-.5\height}{\includegraphics[width=0.15\textwidth]{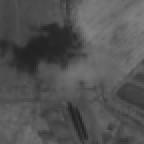}} & \raisebox{-.5\height}{\includegraphics[width=0.15\textwidth]{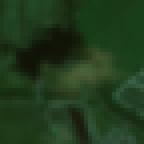}} & \raisebox{-.5\height}{\includegraphics[width=0.15\textwidth]{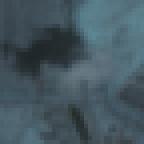}} & \raisebox{-.5\height}{\includegraphics[width=0.15\textwidth]{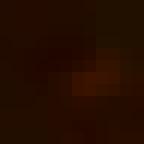}} \\
        Proc Plants & \raisebox{-.5\height}{\includegraphics[width=0.15\textwidth]{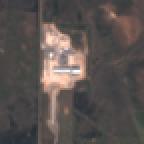}} & \raisebox{-.5\height}{\includegraphics[width=0.15\textwidth]{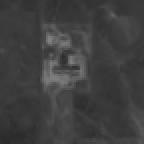}} & \raisebox{-.5\height}{\includegraphics[width=0.15\textwidth]{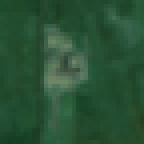}} & \raisebox{-.5\height}{\includegraphics[width=0.15\textwidth]{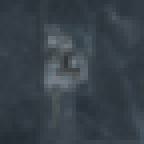}} & \raisebox{-.5\height}{\includegraphics[width=0.15\textwidth]{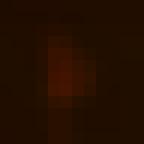}} \\
        R\&Ts & \raisebox{-.5\height}{\includegraphics[width=0.15\textwidth]{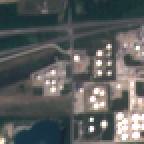}} & \raisebox{-.5\height}{\includegraphics[width=0.15\textwidth]{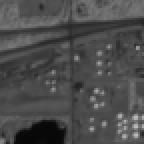}} & \raisebox{-.5\height}{\includegraphics[width=0.15\textwidth]{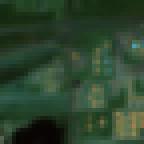}} & \raisebox{-.5\height}{\includegraphics[width=0.15\textwidth]{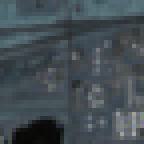}} & \raisebox{-.5\height}{\includegraphics[width=0.15\textwidth]{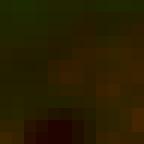}} \\
        WWTPs & \raisebox{-.5\height}{\includegraphics[width=0.15\textwidth]{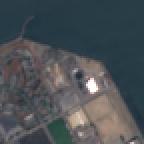}} & \raisebox{-.5\height}{\includegraphics[width=0.15\textwidth]{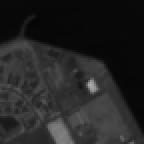}} & \raisebox{-.5\height}{\includegraphics[width=0.15\textwidth]{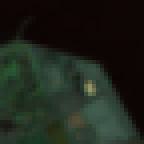}} & \raisebox{-.5\height}{\includegraphics[width=0.15\textwidth]{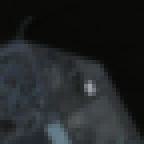}} & \raisebox{-.5\height}{\includegraphics[width=0.15\textwidth]{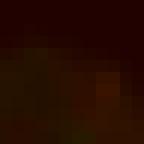}} \\
        Negatives & \raisebox{-.5\height}{\includegraphics[width=0.15\textwidth]{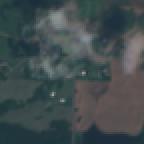}} & \raisebox{-.5\height}{\includegraphics[width=0.15\textwidth]{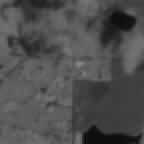}} & \raisebox{-.5\height}{\includegraphics[width=0.15\textwidth]{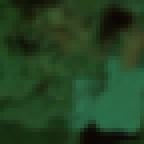}} & \raisebox{-.5\height}{\includegraphics[width=0.15\textwidth]{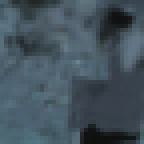}} & \raisebox{-.5\height}{\includegraphics[width=0.15\textwidth]{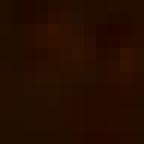}} \\
    \end{tabular}
\end{figure*}

\end{document}